\documentclass{vgtc}                          

\graphicspath{{figures/}{pictures/}{images/}{./}} 

\usepackage{times}                     

\usepackage{tabu}                      
\usepackage{booktabs}                  
\usepackage{lipsum}                    
\usepackage{mwe}                       

\usepackage{mathptmx}                  

\onlineid{0}

\vgtccategory{Research}

\vgtcinsertpkg

\title{“Show Me What’s Wrong!”:\\Combining Charts and Text to Guide Data Analysis}

\author{Beatriz Feliciano\thanks{e-mail: beatriz.feliciano@feedzai.com} %
\and Rita Costa\thanks{e-mail: rita.costa@feedzai.com} %
\and Jean Alves\thanks{e-mail: jean.alves@feedzai.com} %
\and Javier Liébana\thanks{e-mail: javier.liebana@feedzai.com} %
\and Diogo Duarte\thanks{e-mail: diogo.duarte@feedzai.com} %
\and Pedro Bizarro\thanks{e-mail: pedro.bizarro@feedzai.com}}
\affiliation{\scriptsize Feedzai}

\teaser{
  \centering
  \includegraphics[width=\linewidth]{figures/proposed_interface.png}
  \caption{Proposed interface with multiple levels of detail exploration composed of (A) a region where the alert is segmented in the subgroups that compose it (A.1, A.2, A.3, A.4, A.5, and A.6) and where groups that require more attention (in this case, A.5) are highlighted in red; (B) an automatically generated text summary of a selected area (A.3) that provides a broad understanding of the group; and (C) an interactive graphical representation of all the data points of the selected area to explore information in detail.}
  \label{fig:teaser}
}

\abstract{
    Analyzing and finding anomalies in multi-dimensional datasets is a cumbersome but vital task across different domains. In the context of financial fraud detection, analysts must quickly identify suspicious activity among transactional data. This is an iterative process made of complex exploratory tasks such as recognizing patterns, grouping, and comparing. To mitigate the information overload inherent to these steps, we present a tool combining automated information highlights, Large Language Model generated textual insights, and visual analytics, facilitating exploration at different levels of detail. We perform a segmentation of the data per analysis area and visually represent each one, making use of automated visual cues to signal which require more attention. Upon user selection of an area, our system provides textual and graphical summaries. The text, acting as a link between the high-level and detailed views of the chosen segment, allows for a quick understanding of relevant details. A thorough exploration of the data comprising the selection can be done through graphical representations. The feedback gathered in a study performed with seven domain experts suggests our tool effectively supports and guides exploratory analysis, easing the identification of suspicious information.
}


\begin{document}



\firstsection{Introduction}

\maketitle


Data exploration for risk analysis is a crucial task in multiple domains such as financial loan approval~\cite{sheikh2020approach}, medical diagnosis~\cite{alvi2021deep}, and financial fraud detection~\cite{LeiteEVA}. Analysts must, among other tasks, identify data patterns and compare a current event (e.g., loan application, health exam parameters, or performed transaction) with historical information (e.g., previous loans, patient history, or transactions through time). For this, domain experts need to carry out a non-linear and iterative data exploration process~\cite{InteractiveVisualAnalytics} based on tabular data, which makes exploring information and detecting events a complex, fatiguing, and time-consuming task~\cite{MacasATOVis}. 

Although a common problem across domains, our work focuses on the detection of financial fraud -- an issue that resulted in the loss of 159 billion dollars in the USA in 2023~\cite{FeedzaiFinancialLoss}. To better understand how the current analysis process is carried out in this use case, we conducted observation sessions with eleven financial fraud analysts. This enabled us to define design goals tailored to experts' specific needs, which informed our proposal outline.

We observed that, when reviewing a financial fraud alert, analysts come across hundreds or even thousands of data points that they need to explore and analyze to reach a conclusion. Adding to the complexity of the issue, this process must be carried out rapidly~\cite{LeiteEventDetection}, in a few minutes or even seconds~\cite{MacasFraudDetection}, through the exploration of tabular data~\cite{MacasATOVis}. Analysts frequently come across multiple data aggregations and need to switch from the overall view to a more granular exploration of a given high-level concept (e.g., geographical information). These changes in granularity frequently lead to a loss in overall context. Additionally, financial analysts often work alongside automated systems designed to aid them in their decision making~\cite{MacasATOVis} by providing a risk estimate in the form of a risk-score, and even by performing risk-attribution, identifying which variables contribute to an elevated risk-score, if any. This adds further complexity to the task, as analysts are not only required to explore the data themselves but also to incorporate the automated system's insights into their judgment.

To aid in these issues and leverage human perception, current research proposes different solutions, from data summarization -- through graphical representations~\cite{TargetVue, MixedDataExploration, BitConduite, VisPlot.DrillDownFallacies, LeiteEVA, MacasGlyphSOMe, VaBank, VASABI, Medley, Crowdscape, MacasFraudDetection, FraudAuditor} or textual descriptions~\cite{CharatcerizingAutomatedDataInsights, ChartToText, DataTales, VisText} -- to information highlighting~\cite{HighlightUI, ChartAnnotations, TextHighlightingTechniques} and multiple levels of detail exploration~\cite{MTV, VaBank, MacasATOVis, MOOCad, MacasFraudDetection, FluxFlow}. These techniques alone impose some limitations on the data evaluation, such as information overload~\cite{VisualAnalyticsScope}.

We present an innovative system, including a novel user interface promoting time-efficient and highly-informed decision-making. This is achieved by combining a multi-modal summary of the available data in textual and visual representations. Furthermore, the tool integrates the insights from fraud detection methods, leveraging them to produce representations that convey the key risk-assessment insights provided by automated systems. As presented in Figure~\ref{fig:teaser}, the interface is composed of three different elements: (A) a console for guiding analysts by presenting each area of analysis and allowing them to \textbf{detect} which contribute to an elevated risk-assessment; (B) a \textbf{Large Language Model (LLM)} generated text summary containing the relevant information and automated system insights for the selected analysis area, to help them \textbf{understand} its characteristics and why it is particularly risky or not; (C) an interactive graphical representation of all the data points of the selected area that varies and is adjusted according to its attributes, letting users further \textbf{explore} the data. Each interface element (from left to right) has an increased level of detail and can be seen by users at any time, preventing context loss and aiding information retrieval. The text acts as the link between the high-level representation of the data and the more complex and detailed charts.


To evaluate our system we performed a qualitative assessment for a financial fraud detection use case with seven domain experts. The interface was praised for its ease of use and integration into existing workflows, reducing the current analysis workload and allowing a faster insights retrieval. The console (A) and text summary (B) received particular acclaim for their effectiveness in providing concise summaries of alerts and the areas that compose them, thereby eliminating the necessity to examine all data points in simpler cases.

The key contributions of our work are:
\begin{itemize}
    \item Identification of the characteristics of exploratory analysis in the context of a financial fraud use case based on observation sessions conducted with eleven fraud analysts;
    \item An interactive interface tailored for exploring and identifying anomalies in large multi-dimensional financial datasets, guiding the analysis and pointing out relevant areas through automated textual and visual insights comprising different levels of detail, which are displayed side-by-side;
    \item A user study with seven financial fraud detection specialists validating the usefulness of the designed interface.
\end{itemize}

\section{Related Work}

This section provides a brief review of prior research on data exploration and high-risk event identification with a focus on the ease of information retrieval by domain experts.

\subsection{Graphical representation of attributes} \label{relwork:graph}
    Event detection is an analytics problem that requires visual exploration~\cite{VisualAnalyticsFoundations}. Consequently, it is common to graphically represent this multi-level data and its attributes~\cite{LeiteEventDetection, LeiteEVA}. 

    Previous research presents various approaches that facilitate behavior and event detection by visually encoding each relevant data attribute through different graphical representations adjusted to the attribute characteristics and the insights that users are looking for~\cite{LeiteEVA, FraudAuditor, VASABI, Crowdscape}. In the field of fraud detection, systems such as EVA~\cite{LeiteEVA} and FraudAuditor~\cite{FraudAuditor} introduce dashboards that represent multiple relevant data attributes. Despite taking advantage of the human perception system~\cite{LeiteEventDetection}, these approaches present all features simultaneously, which can be overwhelming for analysts. 

    Another frequently researched technique to explore multi-dimensional datasets and gather activity patterns is the use of glyph representations~\cite{TargetVue, BitConduite, MacasGlyphSOMe, VaBank}, which encode various data attributes in a unique visualization. In financial use cases, approaches such as VaBank~\cite{VaBank} do so by individually representing each transaction as an element and encoding each of its features by adding multiple visual representations to it. While these approaches reduce the amount of graphs that users need to evaluate, they increase the visualization complexity, as they require the memorization of multiple elaborated visual encodings. Specifically in the field of fraud detection, where analysts are not accustomed to evaluating data in graphical format, it is essential to design simple visualizations~\cite{LeiteEVA}, as verified in the conducted user study (refer to Section~\ref{sec:user-study}).

    Additionally, research also proposes systems that perform automatic recommendations of graphical representations according to the data attributes~\cite{MixedDataExploration}, subsets of data attributes~\cite{VisPlot.DrillDownFallacies} or even to inputted user intentions~\cite{Medley}. However, these techniques are inadequate for time-constrained use cases, as they require manually editing the generated charts for adequate insight retrieval.

    To address the aforementioned issues, our system balances the amount of displayed information and the visualization complexity, providing relevant insights without overwhelming users.

\subsection{Text summary} \label{sec:textsummary}
    The use of textual descriptions of data and charts is a growing technique used to communicate insights such as outliers, data values, extremes, and trends~\cite{CharatcerizingAutomatedDataInsights}. By reading a text description summarizing the data at hand, users are able to immediately retrieve key takeaways, thus having an early advantage in the analysis~\cite{DataTales}. 

    There are several possible approaches to generating these textual insights. One of those is the usage of LLMs to automatically generate this type of content. There are benefits to this technique, namely in producing content that conveys meaningful information about data and its distribution. Nevertheless, despite its usefulness, these insights are more commonly generated by LLMs, which can output inaccurate information (\textit{hallucinations}~\cite{ChartToText, VisText}). Consequently, the unchecked use of LLMs introduces the possibility that the text summary contains misleading or deceptive conclusions, which, in turn, can harm the analyst's information retrieval. 
    
    To minimize the issues described above, state-of-the-art approaches emphasize the importance of providing users with the necessary means to confirm the validity of the information present in the description~\cite{CharatcerizingAutomatedDataInsights, DataTales} -- for example, by allowing access to tabular or graphical representations of the dataset. Our proposed tool includes a Hallucination Detection system (described in Section~\ref{backend:summary}), to further mitigate the possibility of generating erroneous insights.

\subsection{Multiple levels of detail}
    To avoid data overload and facilitate information digestion, certain techniques propose dividing information into layers and presenting details gradually to users. This allows for an evaluation of summarized~\cite{MTV, MacasATOVis, MacasFraudDetection, VaBank} or grouped~\cite{MOOCad, FluxFlow} data, and for exploration, in full detail, of smaller portions of the information. 

    Current research focuses mainly on the use of this technique combined with visual analytics approaches, but we believe that users would also benefit from high-level insights provided in text format, as these have proved to support initial data exploration~\cite{DataTales}. Our system supports textual insights, which are described in further detail in Sections~\ref{sec:ka-text-summary} and~\ref{backend:summary}, combining them with highlighting and exploration techniques to further empower users.

\subsection{Highlight relevant information}
    Highlighting important information, be it within text~\cite{HighlightUI, TextHighlightingTechniques}, tables~\cite{LeiteEVA}, or through annotations within charts~\cite{ChartAnnotations}, is a way of directing users' attention to specific details. In the context of data exploration, information highlighting can guide the analysis process. For example, in EVA~\cite{LeiteEVA} transactional data concerning fraudulent activity is highlighted in red. However, if the wrong information is brought to attention, there is a potential to mislead analysts or delay their decision-making. As such, when combining highlighting techniques with insights that may contain incorrect information -- like the ones generated by LLMs --, interfaces should be accompanied by an explanation of the highlighting identification process~\cite{CharatcerizingAutomatedDataInsights}. Our proposal tackles this issue by providing a textual explanation -- detailed in Sections~\ref{sec:ka-text-summary} and~\ref{backend:summary} -- which gives users an understanding of the reasons for highlighting while working as a summary that helps to quickly gather the main analysis takeaways.

\section{Design goals}
\label{sec:design-goals}

To derive design objectives for this research, we recruited eleven financial fraud analysts from our network. We conducted one-hour remote observation sessions with each of them, to perceive their data review process. The analysts' work consists of reviewing transactions alerted as possibly fraudulent by an automated system and, in a matter of seconds, assessing whether they correspond to illicit activity. To this end, they iteratively check information regarding the currently flagged transaction and compare it with the alerted entity's past activity, evaluating if they are in line with each other and whether the alert corresponds to an anomalous event. 

We were able to determine that, to perform this assessment, analysts divide data exploration into different subsets, henceforth referred to as \textit{Knowledge Areas} (KAs), -- for example, the alerted person's demographics, past alerted transactions, transactional activity, or the counterpart's demographics -- and explore each area at a time in a non-linear way. By dividing exploration into KAs, experts are able to reduce the laboriousness of the analysis and more easily identify unusual activity.

Currently, the data exploration process is entirely based on tabular data. This requires analysts to invest a significant amount of time navigating through tables to gather insights on relevant patterns and identify suspicious areas, adding complexity to the task.


Based on the insights resulting from the conducted sessions and the literature review, we identified the following design goals:
\begin{itemize}
    \item \textbf{(DG1) Identification of KAs requiring more attention}. Users should be able to easily assess all the different areas that compose the data and to quickly understand which ones might require more careful analysis.
    \item \textbf{(DG2) Summarization of key information within a KA}. Users should easily retrieve the main takeaways of each KA: the mean of the values, trends (ascending or descending), extremes, and differences, among others. 
    \item \textbf{(DG3) Detailed exploration of the attributes of a KA}. All the data related to each KA should be present in the system, allowing users to gather detailed concept-specific information;
    \item \textbf{(DG4) Comparison between an individual data point} (currently alerted transaction) \textbf{and groups of data points} (historic of transactions). Users should be able to compare the currently examined event with all the data points in the dataset.
\end{itemize}


\section{Visualization design}

To structure the interface, we propose a segmentation of the data into KAs. The following KA overviews are considered for financial fraud detection: (A.1) alerted person, (A.2) location, (A.3) flow balance, (A.4) card, (A.5) counterpart, and (A.6) alerted person activity. Even though we have six KAs in our proposal, the tool's design can accommodate up to 12 areas (but, to avoid information overload, we recommend care in the considered number of KAs).

As illustrated in Figure~\ref{fig:teaser}, our proposal is composed of three elements: (A) a Knowledge Area Console, which is the starting point of the analysis, where the alert is segmented and the analyst is guided to the most relevant KAs, (B) an LLM generated text summary for the selected KA, which states the most pertinent information of the chosen area and where its relevance for the analysis can be assessed, and (C) a graphical representation of all the data of each KA, enabling a detailed analysis for specific cases. Since they provide high-level insights of the alert data, the Knowledge Area Console (A) and the text summary (B) elements are grouped in an area called Alert Overview. The graphical representation element (C) is part of an area with the name of the selected KA, as it consists of detailed information about the chosen subset.

Given analysts' proficiency in exploring tabular data~\cite{LeiteEVA}, our tool includes a data table below the presented approach. This is not reflected in the interface images as it corresponds to the current analysis format and does not consist of a novel proposal.

\subsection{Knowledge Area Console (A)}
\label{sec:ka-console}

    The Knowledge Area Console is the starting point of the analysis and where experts can quickly identify the areas of the data that might require more careful analysis (DG1). It consists of a holistic visual depiction of the data where each concept is represented with an icon. There are as many icons as KAs, which are manually designed by experts depending on the use case. Here, users can select the KA for which the tool provides the automated text summary (B) and the graphical representation (C).
    
    Since the alerted person is the focus of the review for the financial fraud analysts, we chose to represent it as the central element of the Console. This KA corresponds to demographic data and insights about past alerts of the alerted person. Other KAs are placed around it, as they all derive from the alerted person's activity.

    \begin{figure}[!htbp]
    \centering
        \begin{minipage}{\columnwidth}
            \includegraphics[width=\columnwidth]{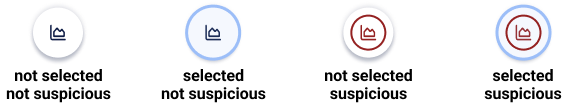}
            \caption{Different states that a KA can have in the Console region.}
            \label{fig:ka-states}
        \end{minipage}
    \end{figure}
    
    As shown in Figure~\ref{fig:ka-states}, KAs have different representations depending on their selection and risk status. We highlight, with a red circle, KAs that are identified as containing potentially suspicious behavior, thus requiring a more careful examination. We describe implementations on how to identify which KAs are suspicious in Section \ref{backend:risk}. This highlight is the first visual cue that communicates to the analyst that something might be out of the ordinary in the data, such that risk analysts interviewed by us claim that, with this design, they would know what is wrong almost immediately.

\subsection{Knowledge Area Text Summary (B)}
\label{sec:ka-text-summary}

    \begin{figure}[!htbp]
    \centering
        \begin{minipage}{\columnwidth}
            \includegraphics[width=\columnwidth]{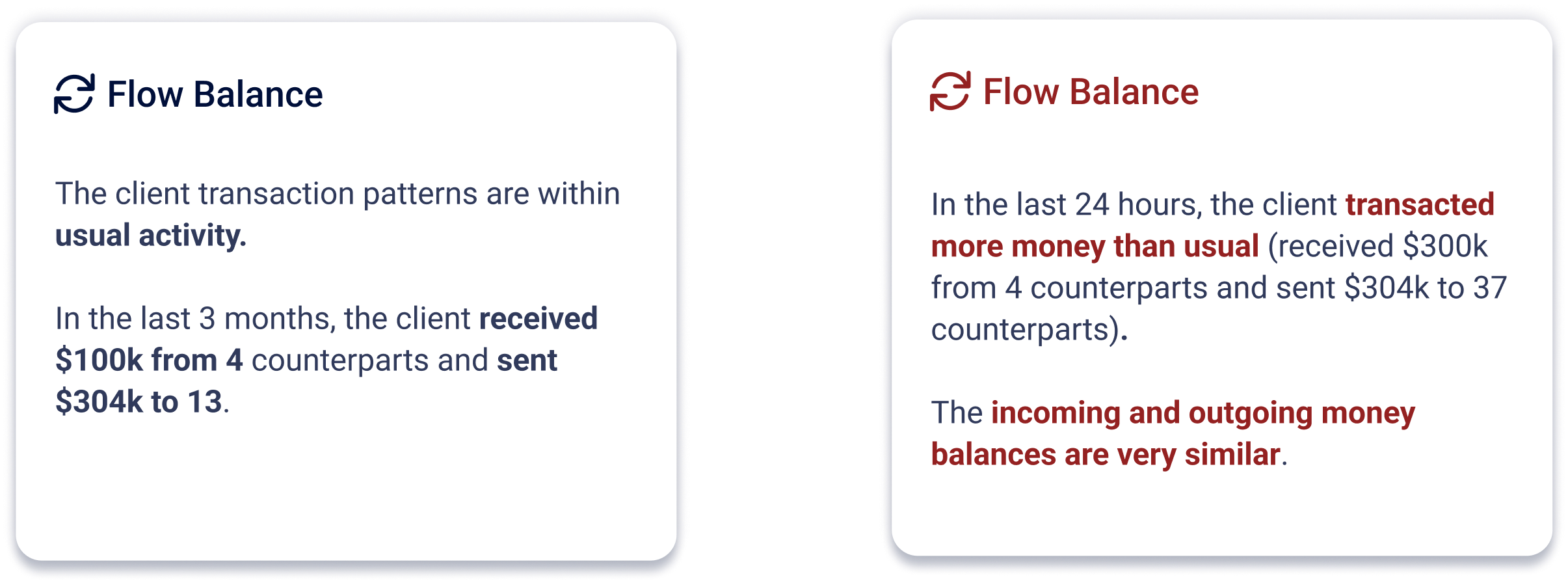}
            \caption{Examples of text summaries for different KAs: alerted person (on the left) and alerted person activity (on the right).}
            \label{fig:ka-summary}
        \end{minipage}
    \end{figure}

    This area serves as the link between the high-level data representation (A), described in Section~\ref{sec:ka-console}, and the detailed insights (C), discussed in Section~\ref{sec:ka-graphs}. Analysts resort to this part of the interface to quickly gather the most relevant insights about the selected KA (DG2) and, in case of identified suspicious information, understand what was marked as risky activity by the system. The main goal of the text summary is to succinctly outline the key values pertaining to the KA, incorporating the insights provided by the automated system. The user study described in Section~\ref{sec:user-study} highlighted that, if the information gathered in this area answers all analysts' questions, no further details are examined and analysis time is saved.
    
    As illustrated in Figure~\ref{fig:ka-summary}, the focus of each KA summary is adjusted not only to the data about the subset but also to the insights that the analysts identified during observation sessions (see Section~\ref{sec:design-goals}) as being most important for the subset in question. Besides stating the main attributes of the area (for the alerted person KA, for example, the person's name, age, and country are represented), it also describes how the current event compares to the alerted entity's behavioral patterns (DG4). Based on those patterns, the system's risk-assessment methods (described in Sections~\ref{backend:risk} and~\ref{backend:summary}) evaluate if the event is an outlier.

    \begin{figure}[!htbp]
    \centering
        \begin{minipage}{\columnwidth}
            \includegraphics[width=\columnwidth]{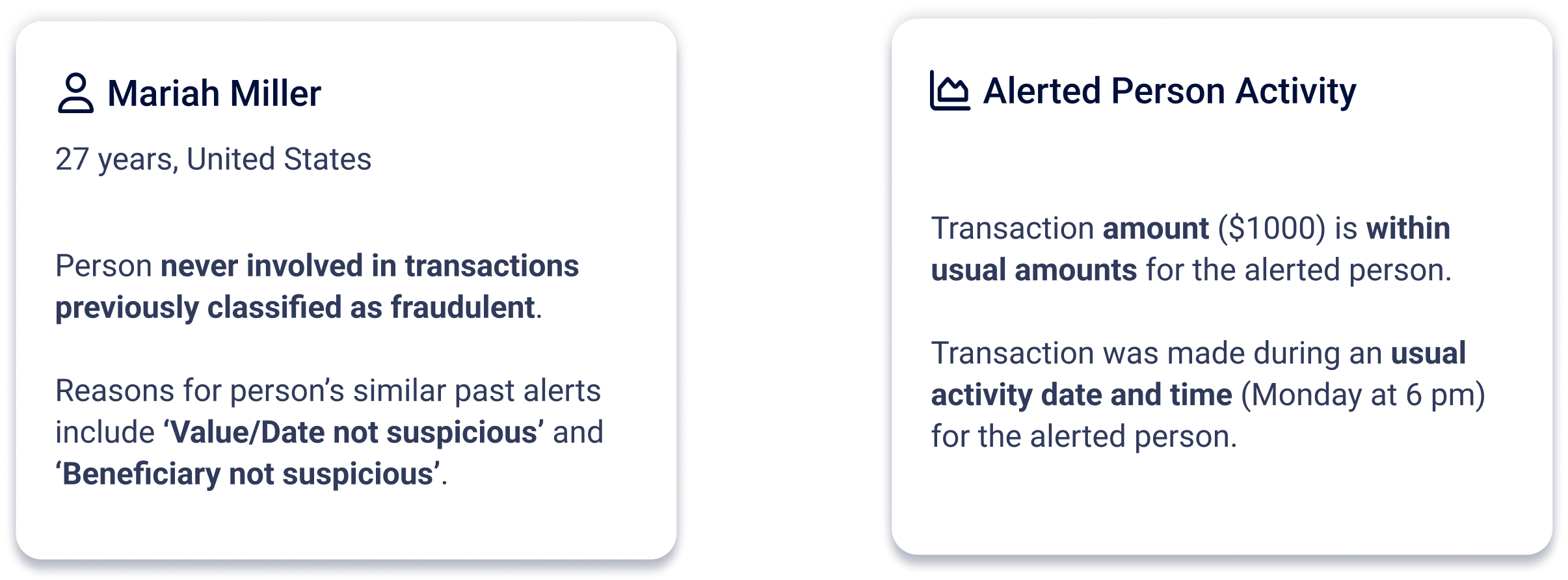}
            \caption{Example of a text summary of a KA not flagged as fraudulent (on the left) and of a KA flagged as fraudulent (on the right).}
            \label{fig:ka-fraud-summary}
        \end{minipage}
    \end{figure}

    Should the summary contain any piece of information indicating that the currently alerted transaction constitutes suspicious activity, that area of the text is highlighted in red, as represented in Figure~\ref{fig:ka-fraud-summary}. Just like the summary, its text highlights -- bold and red color -- are automatically generated using an LLM, described in Section~\ref{backend:summary}.

\subsection{Knowledge Area Graphical Representation (C)}
\label{sec:ka-graphs}

    In this portion of the interface, all the information that composes a KA is graphically represented (DG3). This area is designed for use cases where an exhaustive evaluation of the data is required. It allows analysts to confirm the automated insights provided in the text summary (B) area, as it has been shown that analysts can be critical of erroneous automated insights~\cite{de2020case}. Here, the analyst can visualize sets of graphs designed to represent the historical data of the selected KA. Consider, for instance, the flow balance KA (refer to Figure~\ref{fig:teaser}), where it is important for analysts to gather insights about the total incoming and outgoing amounts, the transactional activity evolution through time, and the counterparts involved in it. Here, their goal is to understand if the balance between incoming and outgoing amounts or involved counterparts is suspicious~\cite{MoneyLaundering}.

    \begin{figure}[!htbp]
    \centering
        \begin{minipage}{\columnwidth}
            \includegraphics[width=\columnwidth]{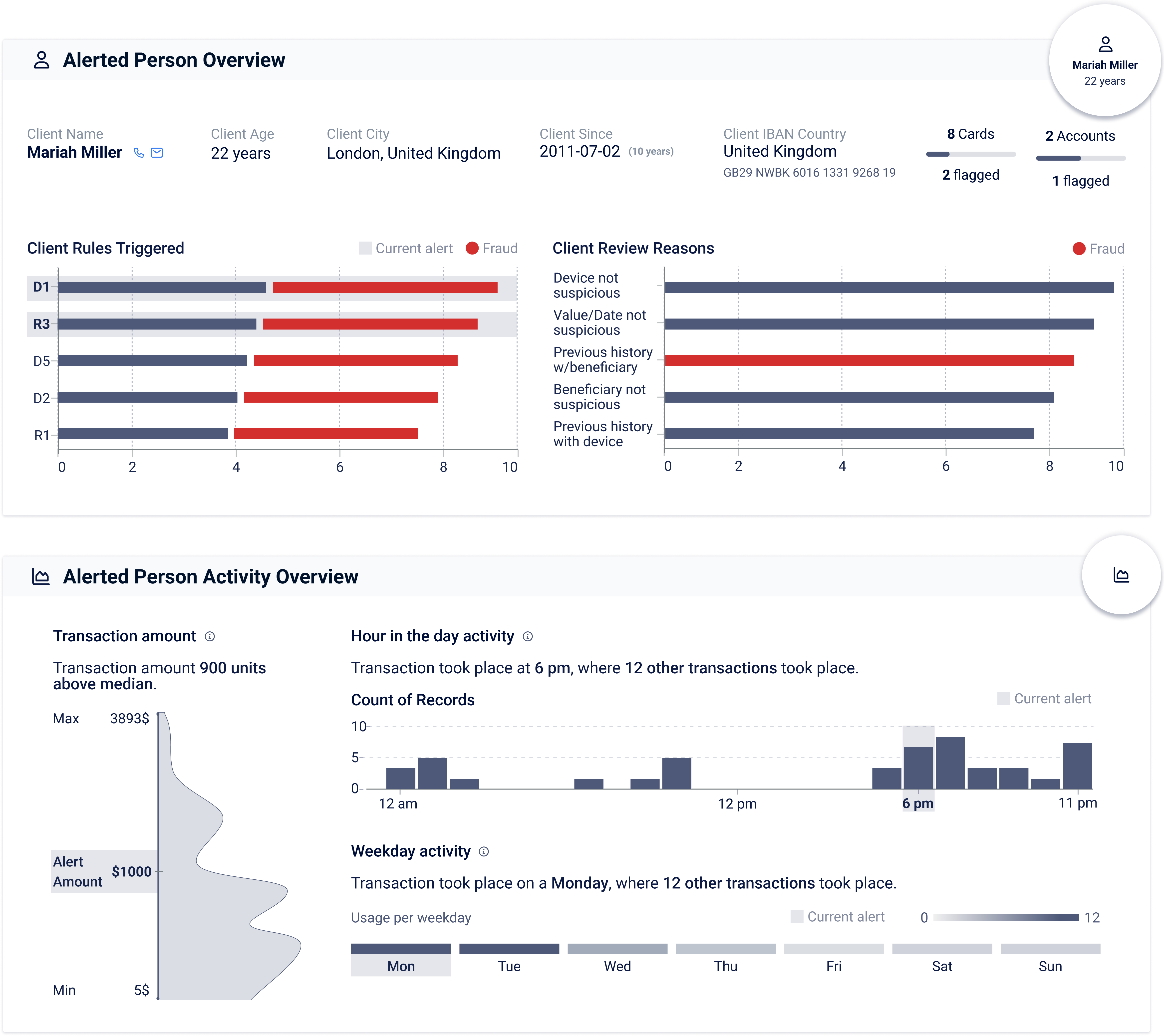}
            \caption{Examples of graphical representations of different KAs: alerted person (on top) and alerted person activity (on the bottom).}
            \label{fig:ka-graphs}
        \end{minipage}
    \end{figure}

    Figure~\ref{fig:ka-graphs} lists some examples of different KAs' graphical representations. Following previous research~\cite{LeiteEVA}, we gave preference to familiar visualizations -- such as bar charts, histograms, stacked bars, and area plots -- to reduce the analysts' learning curve and thus the information retrieval time. Even though we designed a visual depiction for each data subset to leverage the retrieval of insights, we will not cover in further detail their gathered insights and visual encodings,  as these representations are not the focus of this paper.

    Each graphical representation is accompanied by a short subtitle that provides a brief summary of the properties of the graph -- count of elements, mean, max and min values, and so forth. Information about the currently alerted transaction is highlighted in the graphical representations with a gray background, easing its comparison with the depicted history of data (DG4). Similarly, if the subset that constitutes a KA contains data pertaining to a previously confirmed fraudulent event, that information is highlighted in red, as shown in the alerted person overview of Figure~\ref{fig:ka-graphs}.

\section{Backend System}
\label{backend}

\begin{figure*}
    \centering
    \includegraphics[width=\textwidth]{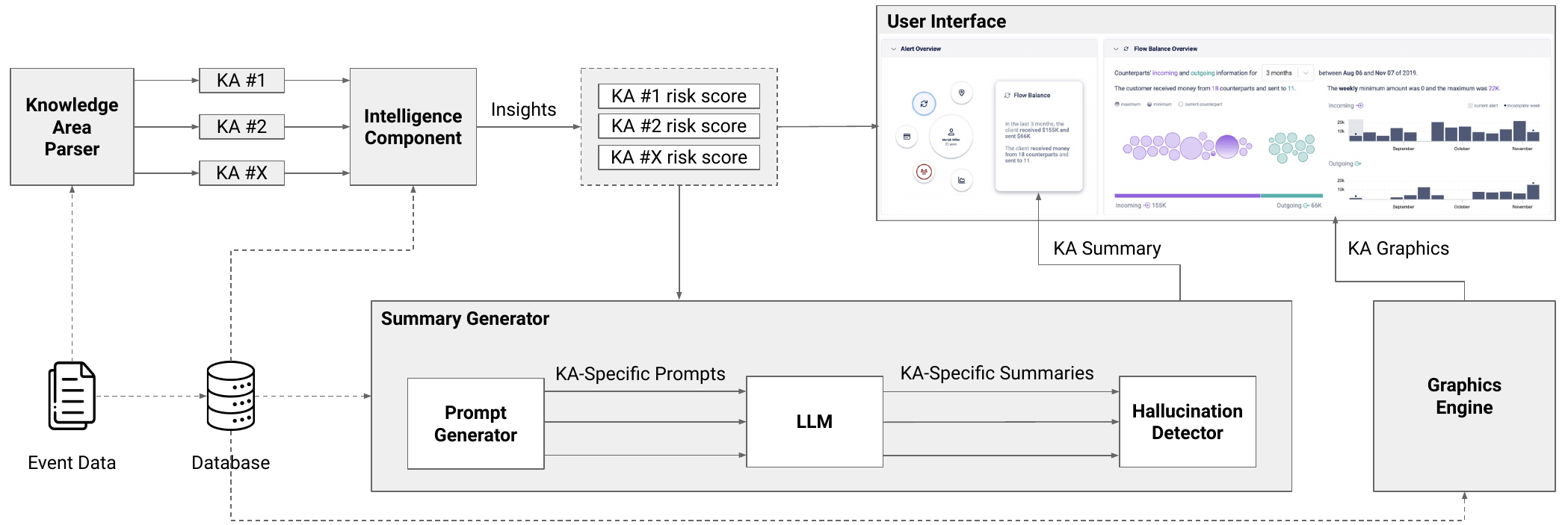}
    \caption{Overview of the architecture of the system.}
        \label{fig:system-architecture}
\end{figure*}

In this section, we describe how the backend system is designed, from the input data to the visual and textual summary generation. In Figure~\ref{fig:system-architecture} we present a schematic representation of the tool’s architecture.

\subsection{Automated Risk-Assessment and Attribution}
\label{backend:risk}

First, the current event data is fed into the system, being stored in the Database component, which contains data from previous transactions. Then, it is fed into the Knowledge Area Parser, which maps the information into KAs. Due to the specific challenges and properties of different use cases (e.g. transaction fraud, bank account opening fraud), domain experts must pre-define each KA and the parsing methods to segment the data into these fields. 

After segmentation, the input data is fed into the Intelligence Component, which produces two outputs for each KA: a risk-score, quantifying the overall risk associated with an event; and, if the score is elevated according to a predefined threshold, a risk-attribution -- an identification of the top-k variables responsible for the elevated risk-score. This information is then used in the Knowledge Area Console (A) representation, described in Section~\ref{sec:ka-console}.


For the current implementation, we utilized an existing rule system comprised of a set of user-defined rules that encode expert knowledge on what constitutes risky behavior. Should a rule be triggered, the KAs associated with the variables that caused said rule to be triggered are flagged as a source of risk. This service may also check a user or device blocklist, in which case, the justification for blocking said entity (i.e., relevant KAs) is saved to the Database at the time of addition to said blocklist. Nevertheless, our approach could be used in combination with other risk assessment methods, such as \textbf{Machine Learning (ML)} models, and in that case we could use well-established attribution explanation methods such as Shapley values~\cite{lundberg2017unified} or LIME~\cite{ribeiro2016should} to identify the key variables responsible for an elevated risk-score.


\subsection{Summary Generation}
\label{backend:summary}

The processed data, along with the outputs of the Intelligence Component, are then fed to the Summary Generator, which is responsible for producing each KA text summary (B), described in Section~\ref{sec:ka-text-summary}. This sub-system follows a Retrieval-Augmented Generation (RAG)~\cite{lewis2020retrieval} approach, comprising three components: (1) The Prompt Generator, responsible for giving proper context, as well as introducing and framing the data and the Intelligence Component insights adequately for use by (2) an LLM, which generates the textual summary based on the provided data and context, and (3) the Hallucination Detector system, which ensures that no factually incorrect data is included in the produced summary.

This structure ensures that, via the generated prompt, all the necessary data and context is fed to the LLM, along with the Intelligence Component's insights (e.g., whether or not a KA is suspicious). 
This approach aims to ensure that the LLM is only tasked with converting the provided information into a human-readable format, in the form of a short paragraph. Similar RAG approaches have been shown to lead to improvement of the generated content while reducing the likelihood of hallucinations~\cite{gao2023retrieval}.

Guaranteeing that the provided summaries are trustworthy is of extreme importance, as a faulty summary could potentially harm the analysts' decision process. For further mitigation of hallucination risk, we include the Hallucination Detector which employs faulty generation detection methods (e.g., SelfCheckGPT~\cite{manakul2023selfcheckgpt}) and, in the case of an identified hallucination, guarantees that sub-par generations are not shown by refusing to provide a summary. For these rare cases, the information \textit{''Summary not available"} is displayed instead. As an alternative solution, we could explore the usage of a dynamic adjustment of the values of generation parameters, for example, by decreasing the value of top-p -- guaranteeing that only the highest probability tokens are sampled -- for the token decoding process. This has been shown to promote factuality in the model's output~\cite{lee2022factuality}, while reducing its variability and increasing the likelihood of repetition in the text generation.

\subsection{Graphics Engine}
\label{sec:graphengine}

The Graphics Engine receives as input all the data concerning the activity of the alerted person to generate a set of format-appropriate visual representations for each KA, described in Section~\ref{sec:ka-graphs}. 

Due to the lack of methods to automatically create visual representations without the need for manual adjustments (see Section~\ref{relwork:graph}), this approach demands a higher level of domain-expert input, requiring graphs to be designed and implemented. After that, each KA graphical representation is fed with its relevant attributes.




\section{User study}
\label{sec:user-study}

To validate the interface, we conducted a user study with seven (P1-P7) of the eleven financial fraud detection analysts who took part in the observation sessions (see Section ~\ref{sec:design-goals}).

\textbf{Data.} We used a dataset with transactional synthetic data, similar to the one proposed by Jesus et al.~\cite{jesusTurningTablesBiased2022}, which was designed to capture the dynamics of fraud in bank account transactions over time, providing a realistic representation of the data encountered in real-world scenarios. The features included in the dataset cover customer information, and transaction details, among others.

\textbf{Procedure.} We conducted one-hour semi-structured interviews with each participant via Zoom. They were briefed about the purpose of the system, how they could interact with it, and its visual encodings. After that, they were given remote access to a computer with the tool (Figure~\ref{fig:teaser}) and were asked to freely interact with it and explore the data for all the KAs while thinking aloud~\cite{ThinkAloud}. They were informed during the sessions that they could ask design or functionality-related questions to the researchers.

\textbf{Findings.} The feedback from the user study proves the usefulness of our tool in aiding data exploration and insight retrieval in multi-dimensional datasets, in the context of financial fraud detection. While P2 and P6 shared some concerns about the automatically highlighted and summarized information potentially introducing bias to the analysis process, all users expressed interest in adopting this tool and emphasized several key benefits of using it.

Experts noted how \textbf{this approach would improve their efficiency}. P1 mentioned that, by looking into the Knowledge Area Console (A) and the subsets' text summary (B), they could, \textit{"in a matter of two seconds, understand what the data is showing"} and added that they were able to classify the alert as fraudulent or not \textit{"almost immediately"}. P3 stated that the multiple levels of detail exploration almost eliminates the need to see data in a tabular format, which would be necessary only in very specific cases where all details need to be considered. Experts believe that, for the vast majority of the cases, the text summary and graphical representation of the data provide all the required granular insights.

Furthermore, P4 highlighted that this tool "would enable the most basic review to be almost automatic, allowing [analysts] to dedicate [their] time and effort to more advanced cases".

Additionally, participants \textbf{praised the tool's capabilities in conveying behavioral patterns}. P6 stated that it enabled them to \textit{"get an overview of the behaviors of the alerted person"} for each analysis area, and P4 emphasized that, with this system, they could more efficiently gather \textit{"context about the alerted entity's activity and transaction"} than they currently do through tabular data.

Users also mentioned that \textbf{the interface effectively pointed to alert subsets that had potentially risky information}. The visual representations -- particularly the use of colors in the Knowledge Area Console (A) to indicate potentially risky activity and the KA text summary (B) to explain the risk classification -- were highly praised. Users noted that their analysis was driven to highlighted areas, which helped them navigate and focus on the relevant data subsets more efficiently. P3 commented that \textit{"it is easier to identify what is wrong or what is not common to the person's activity"}. P6 added that the tool enabled them to \textit{"easily identify risky areas"} and to quickly understand which unusual information they comprised.

Lastly, participants \textbf{appreciated how all the information was accessible in the same location}. Analysts valued having all the necessary information in one place, with the different levels of detail always visible. P2 mentioned that since \textit{"everything is in the same place, there is no need to scroll and fewer clicks are needed"}, as they do not have to constantly order and filter data. P1 added that this would minimize actions not directly related to retrieving information, leaving more time to analyze data and gather insights.

While the goal of this study was to validate the interface as a whole and not the specific graphical representations of each KA (C), we find it important to highlight that analysts had difficulties interpreting and tended to avoid more complex or less known graphs -- such as the bubble chart shown in Figure~\ref{fig:teaser}. This indicates that the detailed visual depiction of some KAs will require adjustments.


\section{Conclusion and discussion}

Detecting anomalous activity in transactional datasets requires navigating and interpreting complex and multi-dimensional data fast and accurately. Despite that, users often suffer from information overload, which leads to performance issues in information retrieval. The proposed tool addresses this issue by dividing data into subsets, highlighting the ones that might require more urgent analysis, and providing a multi-modal summary of the available data per area, both in LLM-generated text and in one or multiple charts. This aids users in simultaneously gathering high-level and granular insights from the data, enabling them to identify patterns, group, and compare information. The performed user study has shown that our solution promotes time-efficient and highly informed decision-making, contributing to reduced work and lower cognitive overload in the fraud analysis process. For future work, we aim to adjust the graphical representation of some analysis subsets and to perform formal user tests to quantitatively measure the contributions of our proposal. Furthermore, we plan on testing the usage of ML models for the risk assessment of the data and further study on hallucination mitigation methods for producing the text summary. 


\acknowledgments{
We want to thank Diana Orghian and Joana Duarte for their support conducting the observation sessions.
We also thank our study participants for their
time and helpful feedback, in special Vanessa Lopes for her continuous insights.
Finally, we thank the anonymous reviewers for their comments.}


\bibliographystyle{abbrv-doi}

\bibliography{template}
\end{document}